\documentclass[letterpaper, 10 pt, conference]{ieeeconf}

\usepackage[utf8]{inputenc}
\usepackage[T1]{fontenc}
\usepackage{graphicx}
\usepackage{amsmath,amssymb}
\usepackage{cite}
\usepackage{booktabs}
\usepackage{pifont}
\newcommand{\cmark}{\ding{51}}
\newcommand{\xmark}{\ding{55}}
\usepackage{placeins}  %
\usepackage{xcolor}
\usepackage{colortbl}
\definecolor{cmpYes}{HTML}{DFF1E1}
\definecolor{cmpNo}{HTML}{F8DEDC}
\definecolor{cmpPart}{HTML}{FBEFC4}
\newcommand{\cyes}{\cellcolor{cmpYes}\textcolor[HTML]{1E7B34}{\cmark}}
\newcommand{\cno}{\cellcolor{cmpNo}\textcolor[HTML]{B3261E}{\xmark}}

\definecolor{dmaDex}{HTML}{1D4ED8}
\definecolor{dmaMimic}{HTML}{C2410C}
\definecolor{dmaAnything}{HTML}{15803D}
\DeclareRobustCommand{\DMA}{\textbf{DMA}}
\DeclareRobustCommand{\DMAColor}{\textbf{\textcolor{dmaDex}{D}\textcolor{dmaMimic}{M}\textcolor{dmaAnything}{A}}}

\DeclareRobustCommand{\DexMimicAnythingColor}{\textbf{\textcolor{dmaDex}{Dex}\textcolor{dmaMimic}{Mimic}\textcolor{dmaAnything}{Anything}}}
\definecolor{realTrack}{HTML}{F2F5F8}
\definecolor{realTrackHead}{HTML}{DEE7EF}
\definecolor{realTele}{HTML}{F7F4F0}
\definecolor{realTeleHead}{HTML}{EDE7DF}
\definecolor{tSeen}{HTML}{F1F7F3}
\definecolor{tUnseen}{HTML}{F7F3FB}
\definecolor{tSeenHead}{HTML}{DCECE2}
\definecolor{tUnseenHead}{HTML}{E8DEF2}
\definecolor{tSeenOurs}{HTML}{DCEDE2}
\definecolor{tUnseenOurs}{HTML}{EADDF5}
\usepackage{tabularx}
\usepackage{multirow}
\makeatletter\newcommand{\setcaptype}[1]{\def\@captype{#1}}\makeatother  %
\usepackage{microtype}  %
\title{\LARGE\bf \DexMimicAnythingColor{}: Learning Generalizable In-Hand Manipulation Skills from Human Demonstrations}

\IEEEoverridecommandlockouts
\makeatletter\let\dma@classmakefntext\@makefntext\makeatother
\usepackage[bottom]{footmisc}
\makeatletter\AtBeginDocument{\let\@makefntext\dma@classmakefntext}\makeatother
\usepackage[colorlinks=true,urlcolor=dmaDex,linkcolor=black,citecolor=black,hyperfootnotes=false]{hyperref}
\IfFileExists{latexml.sty}{\usepackage{latexml}}{\let\iflatexml\iffalse}
\iflatexml
\author{Han Yang \and Jielin Wu \and
Liang Heng \and Xiangrui Jiang \and Siqiao Huang \and Yide Liu \and Guanqi He \and Xiaohan Liu}
\else
\author{Han Yang$^{*}$, Jielin Wu$^{*,\dagger}$, Liang Heng$^{*}$, Xiangrui Jiang,\\
Siqiao Huang, Yide Liu, Guanqi He, and Xiaohan Liu%
\thanks{$^{*}$These authors contributed equally to this work. $^{\dagger}$Project lead.}%
\thanks{All authors are with Wuji Technology.}\\[2pt]
\raisebox{-11.8pt}[\height][\depth]{\fontsize{13}{16}\selectfont \href{https://dexmimicanything.github.io/}{\textbf{dexmimicanything.github.io}}}%
}

\makeatletter
\long\def\@IEEEaftertitletext{%
  \begin{minipage}{\textwidth}
  \centering
  \includegraphics[width=\textwidth]{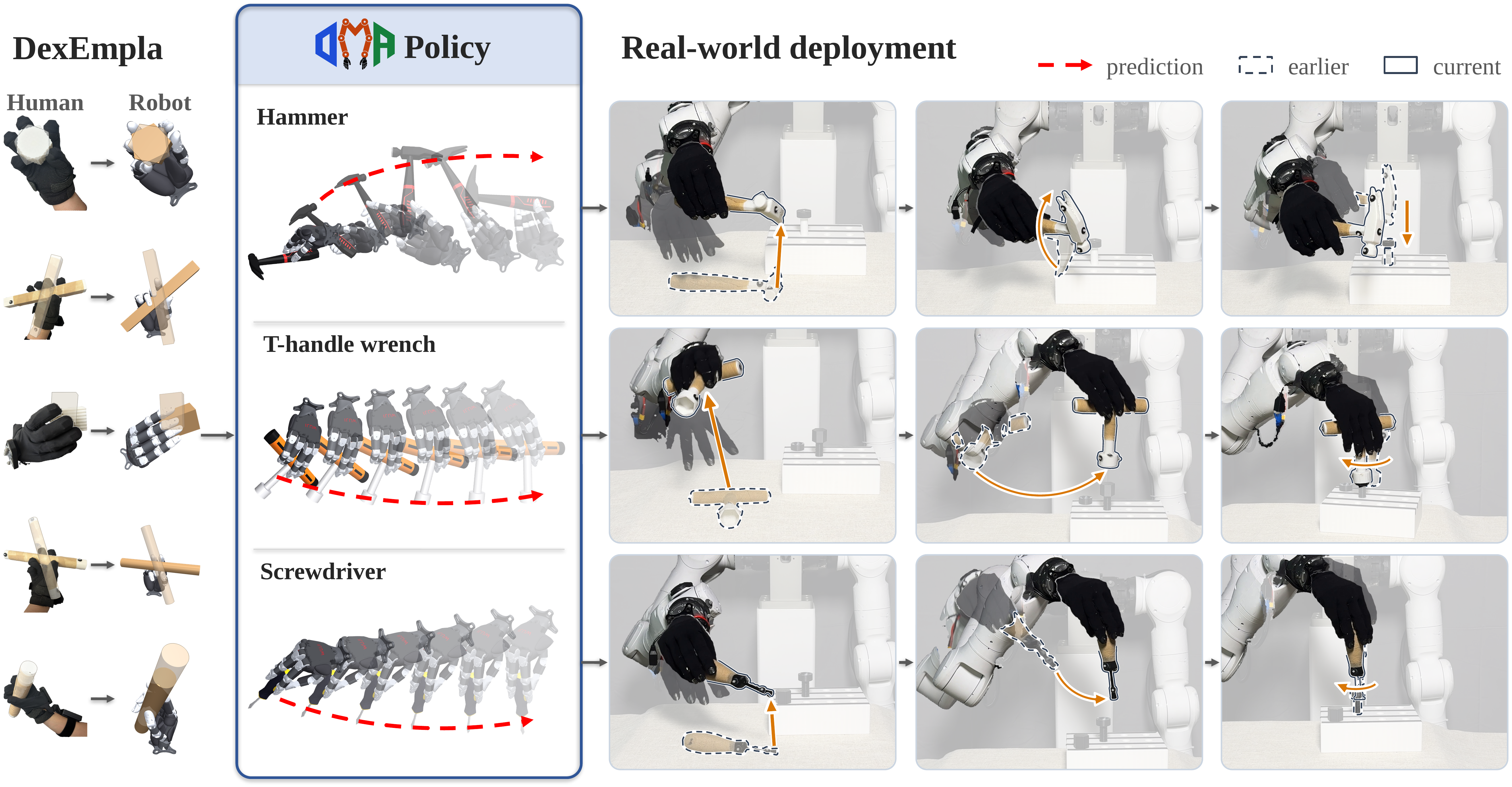}%
  \def\@captype{figure}%
  \caption{\textbf{Learning generalizable in-hand manipulation skills with \DMA{}.} Left:
  DexEmpla provides dexterous hand–object reference trajectories retargeted from human demonstrations.
  Middle:
  Model-based RL trains a single policy to track these references.
  Right: the same policy, trained only in simulation, performs real-world manipulation with unseen tools. In the real-world panels, semi-transparent overlays show earlier poses.}
  \label{fig:teaser}
  \end{minipage}
  \par\vspace{0.5\baselineskip}
}
\makeatother
\fi

\makeatletter
\def\bstctlcite{\@ifnextchar[{\@bstctlcite}{\@bstctlcite[@auxout]}}
\def\@bstctlcite[#1]#2{\@bsphack\@for\@citeb:=#2\do{\edef\@citeb{\expandafter\@iden\@citeb}\if@filesw\immediate\write\csname #1\endcsname{\string\citation{\@citeb}}\fi}\@esphack}
\makeatother

\begin{document}
\bstctlcite{IEEEexample:BSTcontrol}
\maketitle
\iflatexml
\noindent Han Yang, Jielin Wu, and Liang Heng contributed equally to this work.
Jielin Wu is the project lead. All authors are with Wuji Technology.\par
\noindent\href{https://dexmimicanything.github.io/}{\textbf{dexmimicanything.github.io}}\par
\begin{figure*}
  \centering
  \includegraphics[width=\textwidth]{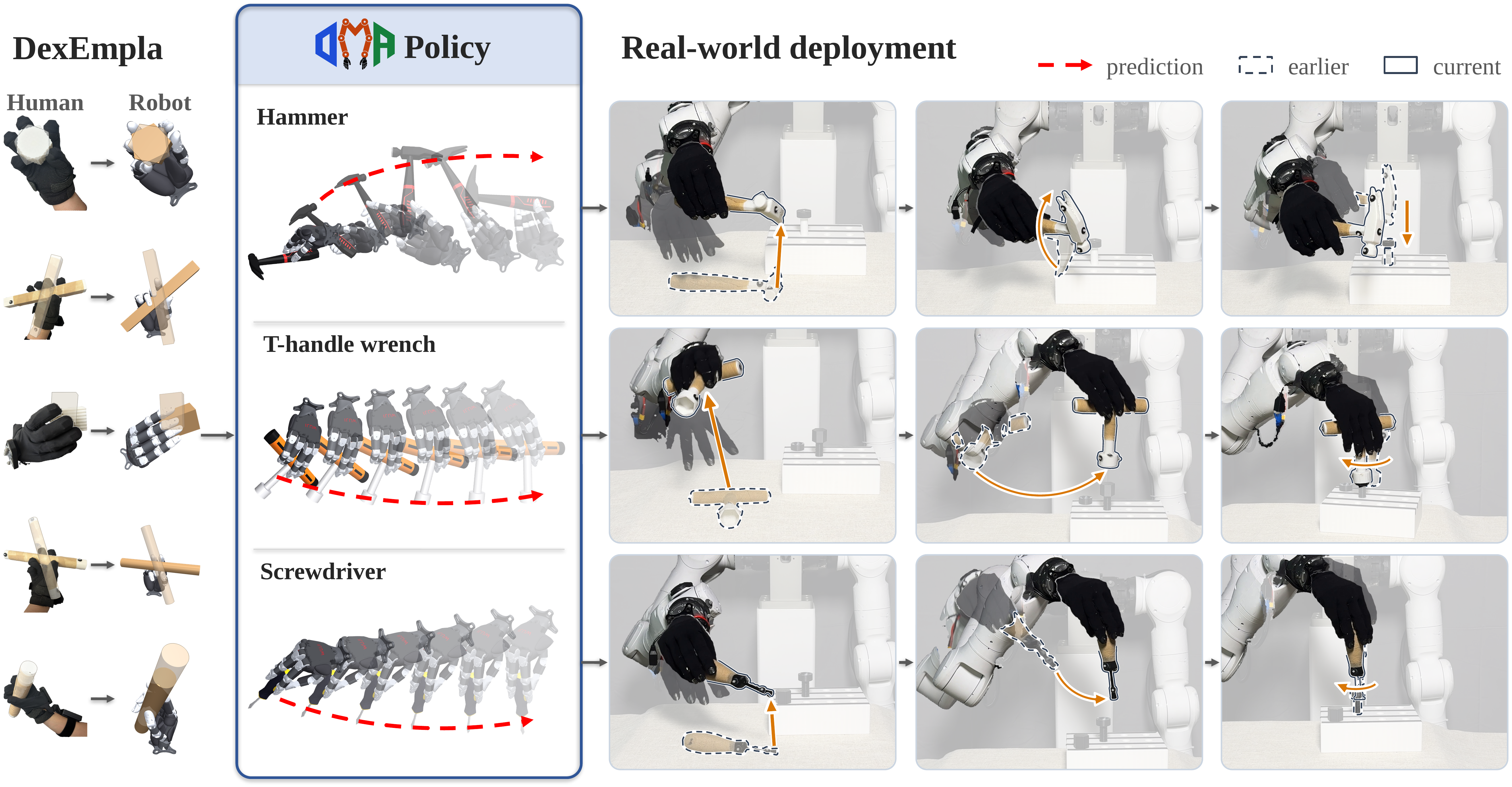}
  \caption{\textbf{Learning generalizable in-hand manipulation skills with \DMA{}.} Left:
  DexEmpla provides dexterous hand–object reference trajectories retargeted from human demonstrations.
  Middle:
  Model-based RL trains a single policy to track these references.
  Right: the same policy, trained only in simulation, performs real-world manipulation with unseen tools. In the real-world panels, semi-transparent overlays show earlier poses.}
  \label{fig:teaser}
\end{figure*}
\fi
\thispagestyle{empty}
\pagestyle{empty}
\begin{abstract}
Humans skillfully use diverse tools, seamlessly transitioning between in-hand manipulation skills such as rotation, translation, and regrasping. However, learning a dexterous manipulation policy that tracks diverse hand–object references across objects and in-hand skills remains challenging.
We present \DexMimicAnythingColor{} (\DMAColor{}), to our knowledge the first hand--object co-tracking controller that learns in-hand skills from human demonstrations and generalizes these skills to new objects and motions without per-object or per-reference training.
To learn a general controller, we build DexEmpla from 40.1 hours of human demonstrations with objects of varied shapes and use model-based reinforcement learning to train a single policy to track the retargeted hand--object motions.
We evaluate \DMA{} in simulation and on hardware.
In simulation, \DMA{} leads the evaluated methods with 80.4\% success on unseen motions of training objects and 69.7\% on novel objects, versus 49.8\% and 32.7\% for the strongest baseline, which trains one policy per test motion.
The same controller completes six tool-use tasks with unseen tools on a Wuji Hand~2, with 71.1\% success in motion tracking and 80.0\% in teleoperation.
\end{abstract}

\section{Introduction}

Humans can use a wide range of tools by flexibly combining in-hand rotation, translation, and regrasping.
For example, using a screwdriver involves adjusting its pose within the hand to align its tip with a screw, then turning it to tighten the screw~\cite{bullock2013,dexscrew}.
Reproducing these skills through teleoperation requires precise finger coordination to maintain object stability as contacts change.
Retargeting errors, limited tactile feedback, and latency make such coordination difficult, hindering the collection of reliable demonstrations~\cite{simtoolreal,penspin,teledexter}.
Even when a successful demonstration is available, open-loop replay cannot correct for deviations caused by small execution errors or disturbances, which may destabilize the object in hand.
Therefore, a robust reactive controller is needed to track the desired object pose in closed loop, by adjusting finger actions based on the current hand--object state.

Reactive hand--object co-tracking controllers~\cite{dextrack,teledexter} are typically trained using synchronized hand and object motion references from human hand--object interaction (HOI) demonstrations.
Policies specialized to individual objects or reference motions can perform in-hand manipulation~\cite{teledexter,adadex} but require additional training for new objects.
Recent generalist trackers~\cite{dextrack,dexmimic,oath} extend to unseen motions and objects, yet their demonstrated capabilities largely emphasize grasping and transport, with more limited evidence of reliable finger-driven rotation, translation, and regrasping across unfamiliar object geometries (Table~\ref{tab:compare}).
Learning a general tracking controller that reuses these in-hand skills across both new objects and reference motions remains challenging.

We present \DexMimicAnythingColor{} (\DMAColor{}), a hand--object co-tracking framework for generalizable in-hand manipulation (Fig.~\ref{fig:teaser}).
Our central idea is to learn reusable in-hand skills from human demonstrations on varied geometric primitives.
We therefore build \textbf{DexEmpla}, an HOI dataset designed specifically for in-hand manipulation, and train a single reactive controller on it through model-based reinforcement learning (MBRL) with a local object representation.
Our contributions are:\looseness=-1
\begin{enumerate}
  \item \textbf{A generalizable hand--object co-tracking controller.} It performs in-hand rotation, translation, and regrasping on unseen objects and references, and combines these skills as each reference requires, without additional training.
  \item \textbf{DexEmpla, an in-hand HOI dataset.} It provides per-frame interaction labels and evaluation references on seen and unseen objects; we will release it.
  \item \textbf{Extensive simulation and real-world evaluation.} We evaluate seen and unseen objects in simulation and motion tracking and teleoperation on hardware, and ablate each component and the training data size.
\end{enumerate}

\section{Related Work}
\begin{table}[t]
  \centering
  \caption{Comparison with representative HOI tracking methods.}
  \label{tab:compare}
  \setlength{\tabcolsep}{1pt}
  \footnotesize
  \renewcommand{\arraystretch}{0.9}
  \begin{tabularx}{\columnwidth}{@{}l*{4}{>{\centering\arraybackslash}X}@{}}
    \toprule
     & Generalist & Unseen & \multicolumn{2}{c}{In-hand evaluation} \\
    \noalign{\vskip-2pt}\cmidrule(lr){4-5}\noalign{\vskip-1.5pt}
    Method & policy & objects & Sim. & Real \\
    \midrule
    ManipTrans~\cite{maniptrans} & \cno & \cno & \cno & \cno \\
    ConTrack~\cite{contrack} & \cno & \cno & \cyes & \cno \\
    TeleDexter~\cite{teledexter} & \cno & \cno & \cyes & \cyes \\
    DexTrack~\cite{dextrack} & \cyes & \cyes & \cno & \cno \\
    DexMimic~\cite{dexmimic} & \cyes & \cyes & \cno & \cno \\
    OATH~\cite{oath} & \cyes & \cyes & \cno & \cno \\
    HOT~\cite{hot} & \cyes & \cyes & \cyes & \cno \\
    \midrule
    \textbf{\DMAColor{} (ours)} & \cyes & \cyes & \cyes & \cyes \\
    \bottomrule
  \end{tabularx}
  \par\smallskip
  {\raggedright\footnotesize \textcolor[HTML]{1E7B34}{\cmark}: quantitative; \textcolor[HTML]{B3261E}{\xmark}: qualitative or none. In-hand: the object is rotated or translated within the hand.\par}
\end{table}
\subsection{Object-Centric RL for In-Hand Manipulation}
Object-centric RL specifies in-hand tasks using an object rotation axis, a target pose, or a trajectory, without providing a hand-motion reference.
Single-skill policies learn in-hand rotation~\cite{hora} or screwing~\cite{dexscrew} with rewards and environments designed for that skill.
More general formulations condition one policy on object goals: UniCross~\cite{unicross} specifies target object states, while SimToolReal~\cite{simtoolreal} and Dex4D~\cite{dex4d} follow object trajectories and point tracks.
Although these policies generalize across objects from object goals alone, the same object motion needs different finger postures and contact schedules on different geometries.
RL has to find both by exploration, which remains a bottleneck even at scale~\cite{simtoolreal} and rarely discovers finger gaiting or regrasping~\cite{adadex}.
\DMA{} conditions on synchronized human hand--object references, which give the finger motion through each contact switch.

\subsection{Hand--Object Tracking from Human References}
Hand--object tracking methods follow a synchronized hand and object reference built from a human demonstration (Table~\ref{tab:compare}).
ManipTrans~\cite{maniptrans} and ConTrack~\cite{contrack} train one policy per reference, and SPIDER~\cite{spider} tracks each reference by sampling-based optimization.
Trackers specialized to individual objects perform in-hand skills under finger contact changes, but each new object needs new data and training~\cite{teledexter,adadex}.

Generalist trackers such as DexTrack~\cite{dextrack}, DexMimic~\cite{dexmimic}, OATH~\cite{oath}, and Dexplore~\cite{dexplore} train mostly on references from public HOI datasets such as~\cite{grab,taco}, which rarely focus on in-hand manipulation.
Methods using synthetic references learn rotation and regrasping only through skill-specific teachers~\cite{hot}.

\DMA{} trains a single controller on \textbf{DexEmpla}, a set of in-hand demonstrations on geometric primitives, and executes varied in-hand skills on unseen objects and references without retraining.

\subsection{Model-Based RL with Learned World Models}
Model-based RL learns a model of the dynamics and uses it to train the policy~\cite{dreamer} or to plan~\cite{tdmpc2,pddm}.
In motion tracking, the references differ, but they all share the dynamics of the body and of contact.
Model-free RL does not model these shared dynamics and keeps re-estimating the effect of actions from rollouts.
SuperMimic~\cite{supermimic} learns them once, as a single world model for a whole motion library.
Its world model covers the robot body only, with no object interaction.
Objects differ in shape, yet the local surfaces that fingers touch are similar~\cite{dexrepnet,demomimic}.
For in-hand rotation, a world model learned on nine objects was reused for 49~\cite{wmcraftnet}.
\DMA{} therefore learns one hand--object world model for all objects and references.
The model also predicts multi-finger contact, which decides when its rollouts reward object tracking.

\FloatBarrier
\section{Method}
\suppressfloats[t]
\begin{figure*}[t]
  \centering
  \includegraphics[width=\textwidth]{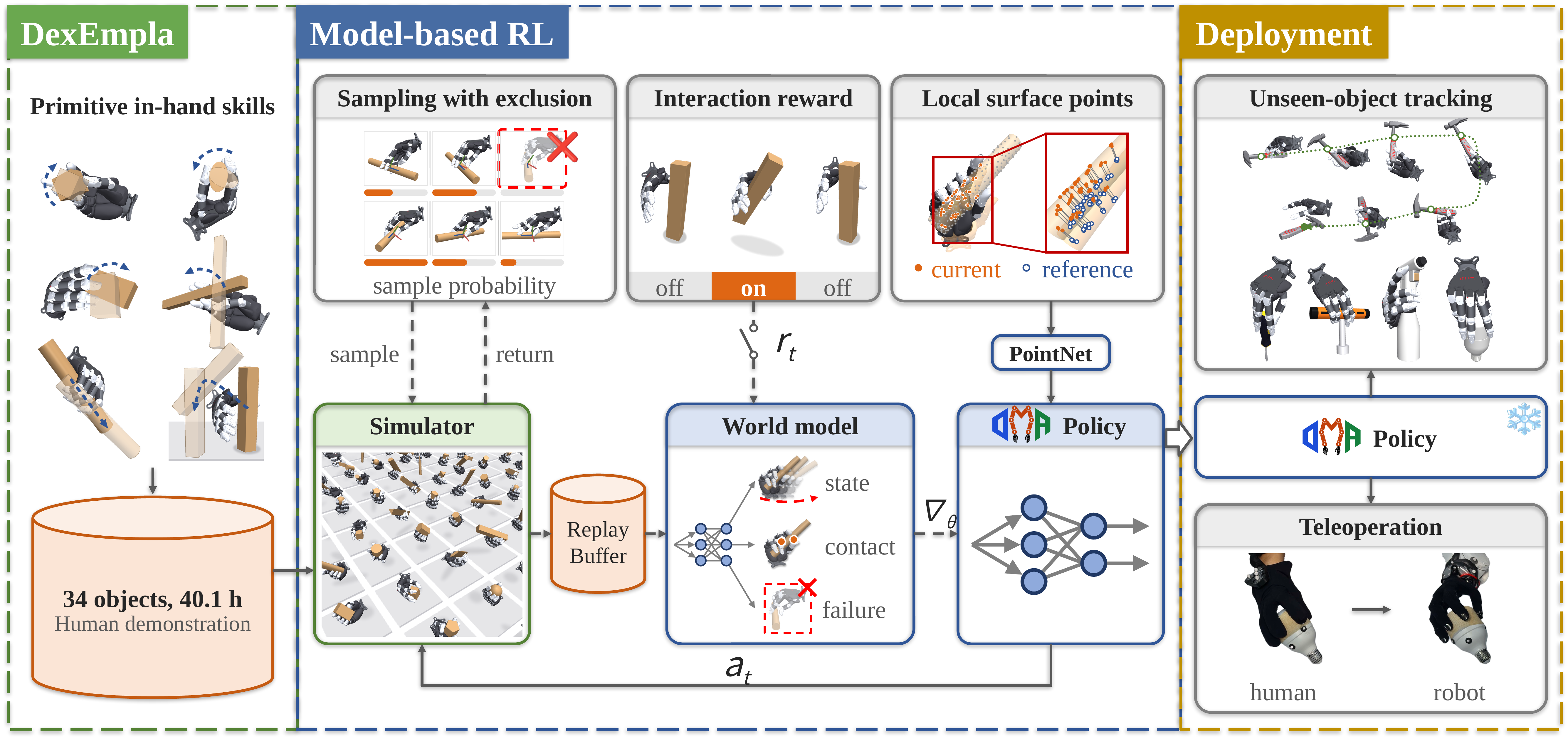}
  \caption{\textbf{Overview of \DMA{}.} (a) DexEmpla provides retargeted hand--object references and interaction labels. (b) Model-based RL with a differentiable world model~\cite{supermimic} trains a shared policy using local surface point pairs, interaction-aware object tracking, and return-based reset sampling that excludes persistently failing starts. (c) The policy is deployed without fine-tuning for motion tracking and teleoperation.}
\label{fig:overview}
\end{figure*}

\subsection{Problem Formulation}
\label{sec:formulation}
We formulate hand--object co-tracking as goal-conditioned control (Fig.~\ref{fig:overview}).
The inputs are an object mesh and a reference $\xi=\{\bar T_{w,k},\bar T_{o,k},\bar q_k,\bar P_k,h_k\}_{k=0}^{N-1}$, where $\bar T_{w,k},\bar T_{o,k}\in\mathrm{SE}(3)$ are the wrist and object poses in the world frame, $\bar q_k\in\mathbb{R}^{20}$ the finger joint angles, $\bar P_k\in\mathbb{R}^{21\times 3}$ the hand link positions, and $h_k\in\{0,1\}$ an interaction label that is 1 when the hand must support the object.
A bar marks a reference quantity, a tilde a noisy observation, and a hat a world-model prediction.
We seek a policy $\pi_\theta$ that tracks the object's reference pose while the wrist follows the reference.

\textbf{Action.}
Human wrist motion is less affected by the embodiment gap than finger motion~\cite{objdex}, so the 6-DoF floating wrist follows $\bar T_{w,k}$ directly and the policy controls the 20 finger joints.
At control step $t$, the reference is at frame $k_t$.
The policy $a_t=\pi_\theta(o_t,g_t)$ outputs residual finger joint targets relative to $\bar q_{k_t}$.

\textbf{Observation and goal.}
The observation is $o_t=[q_t,\dot q_t,a_{t-1},z_t]$, where $q_t$ and $\dot q_t$ are the finger joint angles and velocities and $z_t$ is the surface-point feature.
The goal $g_t=[\bar q_{k_t+1},\bar K_{k_t+1},\bar P_{k_t+1}]$ contains the next-frame finger joint angles, six wrist-frame object keypoints $\bar K$~\cite{toporetarget}, and link positions.

\subsection{DexEmpla Dataset}
\label{sec:data}
\begin{table}[t]
\caption{DexEmpla statistics.}
\label{tab:data}
\centering
\footnotesize
\renewcommand{\arraystretch}{0.9}
\setlength{\tabcolsep}{2.5pt}
\begin{tabular}{@{}lrrrrrrr@{}}
\toprule
 & & \multicolumn{3}{c}{Train} & \multicolumn{3}{c}{Eval} \\
\cmidrule(lr){3-5}\cmidrule(l){6-8}
 & Length (mm) & Obj. & Traj. & Hours & Obj. & Seen & Unseen \\
\midrule
Rotation & & & 1435 & 12.1 & & 81 & 42 \\
Translation & & & 418 & 3.7 & & 24 & 17 \\
Combination & & & 3188 & 24.3 & & 210 & 61 \\
\midrule
Compact objects & 42--90 & 15 & 2113 & 15.8 & 15 & 169 & -- \\
Elongated bars & 175--250 & 19 & 2928 & 24.3 & 19 & 146 & -- \\
Unseen objects & 90--256 & -- & -- & -- & 9 & -- & 120 \\
\midrule
Total & 42--256 & 34 & 5041 & 40.1 & 43 & 315 & 120 \\
\bottomrule
\end{tabular}
\par\smallskip
{\raggedright\footnotesize Rotation: a single rotation about one hand axis; Translation: translation only; Combination: all others. Length: longest object dimension. Seen/Unseen: evaluation trajectories on seen and unseen objects.\par}
\end{table}
Few public datasets~\cite{grab,arctic,taco} focus on in-hand manipulation, and it is unclear what demonstrations such a dataset should contain. To build DexEmpla, we organized data collection around how the object moves relative to the wrist, recording translations and rotations both individually and in combination.

\textbf{Collection protocol.}
DexEmpla contains one-handed human demonstrations in which a rigid object is moved within the hand (Table~\ref{tab:data}).
Following~\cite{bullock2013}, the motions are translation along and rotation about each axis of a frame attached to the hand.
They are performed alone, in sequence, or simultaneously.
Demonstrations combine these in-hand motions with regrasping and pick-and-place.
Each motion is recorded with the palm in different orientations relative to gravity.
For each demonstration, the protocol specifies how the object is first grasped and how it should move; operators choose how to move their fingers.

\textbf{Objects and splits.}
The 34 training objects are geometric primitives, mostly prisms and cylinders with different sizes, from thin plates to 250~mm bars.
To test generalization, we hold out trajectories of each training object.
For combination demonstrations, we manually verified that all held-out references are skill combinations unseen in training.
We also record evaluation trajectories on 9 unseen objects: non-convex tools and everyday objects such as hammers, a screwdriver, a spray bottle, and light bulbs.

\textbf{Capture and retargeting.}
A Wuji Glove gives 21 hand keypoints in the wrist frame, and optical motion capture tracks the wrist and object through rigidly attached markers.
Both streams are synchronized, resampled to 50\,Hz, and retargeted to the robot hand in real time~\cite{toporetarget}.

\textbf{Interaction labels.}
The label $h_k$ indicates whether the hand must support the object at frame $k$.
We obtain it with a gravity test in simulation.
The object is released from rest at $\bar T_{o,k}$ with only environmental support.
If it moves, the hand must support it:
\begin{equation}
h_k=\mathbb{I}\big[\,\|\Delta p_k\|>\delta_p\ \lor\ \angle\Delta R_k>\delta_R\,\big],
\label{eq:label}
\end{equation}
where $\Delta p_k$ and $\Delta R_k$ are its displacement and rotation after 0.5\,s (thresholds in Appendix~\ref{app:method}).

\subsection{Hand-Guided Local Object Surface Encoding}
\label{sec:repr}
The object representation strongly affects how a controller shared across objects transfers to new shapes.
Global representations describe the object as a whole, through bounding-box keypoints~\cite{simtoolreal}, basis point sets~\cite{maniptrans,adadex}, point-cloud features~\cite{dextrack}, or current--target point pairs~\cite{dex4d}.
However, such representations may have little in common across objects whose shapes differ widely.
The surface near the hand can be more similar across objects, for example along tool handles~\cite{dexrepnet}.
Hand-centric features describe this surface by the point nearest to each hand keypoint~\cite{dexrepnet,unicross,dexplore}.
\DMA{} instead uses the hand links only to select surface points, near both the current and the reference hand.
The policy receives each selected point's current and reference positions, a local surface point pair (Fig.~\ref{fig:overview}(b)).
We arrived at this design empirically; Sec.~\ref{sec:exp_ablation} compares it with four alternatives.

Let $\mathcal{B}_o=\{b_i\}_{i=1}^{N_b}$ be surface points pre-sampled from the object mesh uniformly at random over its surface area, expressed in the object frame.
The anchor set $\mathcal{A}_t$ holds the 21 link positions of the current hand, expressed in this frame with the observed object pose $\tilde T_{o,t}$.
The anchor set $\bar{\mathcal{A}}_t$ holds those of the reference hand, expressed with the reference object pose $\bar T_{o,k_t}$.
For an anchor set $\mathcal{A}$, let $\rho_a(i)$ be the rank of $b_i$ by distance to anchor $a$, and $\rho_{\mathcal{A}}(i)=\min_{a\in\mathcal{A}}\rho_a(i)$.
At every control step, we select the $K$ points of lowest rank for the current hand and $K$ further points for the reference hand:
\begin{equation}
\mathcal{S}^{c}_t=\operatorname*{arg\,min^{\mathit K}}_{i}\ \rho_{\mathcal{A}_t}(i),\quad
\mathcal{S}^{r}_t=\operatorname*{arg\,min^{\mathit K}}_{i\notin\mathcal{S}^{c}_t}\ \rho_{\bar{\mathcal{A}}_t}(i),
\label{eq:select}
\end{equation}
and $\mathcal{S}_t=\mathcal{S}^{c}_t\cup\mathcal{S}^{r}_t$ ($N_b$ and $K$ in Appendix~\ref{app:method}).
Ranking per link lets every link contribute its nearest point before any contributes a second, so the points cover the surface around the whole hand.
The selection needs no contact detection or reference contact locations.
For each $i\in\mathcal{S}_t$, the current position in the current wrist frame and the reference position in the reference wrist frame are
\begin{equation}
p_{t,i}=T_{w,t}^{-1}\,\tilde T_{o,t}\,b_i,\quad
\bar p_{t,i}=\bar T_{w,k_t}^{-1}\,\bar T_{o,k_t}\,b_i.
\label{eq:local}
\end{equation}
A point encoder~\cite{pointnet} aggregates the $2K$ pairs into the feature $z_t=\mathrm{PointNet}\big(\{[p_{t,i};\bar p_{t,i}]\}_{i\in\mathcal{S}_t}\big)$.
Each pair gives the positions of one surface point in the two wrist frames, without assigning the point to a finger.
How to move each finger given these pairs is left to the policy.
\looseness=-1

\subsection{Interaction-Aware Object Tracking}
\label{sec:reward}
Before a grasp and after a release, the object rests on the environment and matches its reference untouched.
Prior methods activate object or contact rewards by a hand--object distance threshold~\cite{maniptrans} or once the object has been lifted~\cite{simtoolreal}.
However, the hand is within such a distance before it supports the object, and a reward activated once stays active after a release.
\DMA{} instead rewards object tracking only when the reference requires hand support and the hand holds the object.
Let $n^{c}_t$ be the number of fingers with any link in contact with the object, excluding the palm.
Following contact-gated rewards~\cite{dexpoint}, the gate is
\begin{equation}
\chi_t=h_{k_t}\,\mathbb{I}[n^{c}_t\ge 2].
\label{eq:gate}
\end{equation}
The reward is
\begin{equation}
r_t=\Delta t\big[\chi_t\,r^{\mathrm{obj}}_t+r^{\mathrm{hand}}_t-r^{\mathrm{reg}}_t\big],
\label{eq:reward}
\end{equation}
where $\Delta t$ is the control period.
Appendix~\ref{app:method} defines the object, hand, and regularization terms and lists the weights and kernel widths.
The gate removes the object term on frames in which the environment supports the object and requires two-finger contact on the other frames.
On frames with $h_{k_t}=0$, the joint-tracking weight in the hand term rises, which supervises approach and release.

\begin{table*}[!t]
\centering
\caption{Comparison with baselines on the DexEmpla evaluation references.}
\label{tab:sim}
\setlength{\tabcolsep}{2.5pt}
\renewcommand{\arraystretch}{1.3}
\small
\setlength{\heavyrulewidth}{0.4pt}\setlength{\lightrulewidth}{0.4pt}\setlength{\cmidrulewidth}{0.4pt}\setlength{\arrayrulewidth}{0.4pt}
\setlength{\aboverulesep}{0pt}\setlength{\belowrulesep}{0pt}
\begin{tabularx}{\textwidth}{@{}>{\hsize=2.7\hsize\centering\arraybackslash}X>{\columncolor{tSeen}\hsize=1.05\hsize\centering\arraybackslash}X>{\columncolor{tSeen}\hsize=.775\hsize\centering\arraybackslash}X>{\columncolor{tSeen}\hsize=.775\hsize\centering\arraybackslash}X>{\columncolor{tSeen}\hsize=.775\hsize\centering\arraybackslash}X>{\columncolor{tSeen}\hsize=.775\hsize\centering\arraybackslash}X>{\columncolor{tUnseen}\hsize=1.05\hsize\centering\arraybackslash}X>{\columncolor{tUnseen}\hsize=.775\hsize\centering\arraybackslash}X>{\columncolor{tUnseen}\hsize=.775\hsize\centering\arraybackslash}X>{\columncolor{tUnseen}\hsize=.775\hsize\centering\arraybackslash}X>{\columncolor{tUnseen}[\tabcolsep][0pt]\hsize=.775\hsize\centering\arraybackslash}X@{}}
\toprule
 & \multicolumn{5}{>{\columncolor{tSeenHead}}c}{Seen objects} & \multicolumn{5}{>{\columncolor{tUnseenHead}}c}{Unseen objects} \\
\cline{2-11}
 & \cellcolor{tSeenHead} & \multicolumn{2}{>{\columncolor{tSeenHead}}c}{\iflatexml Object Tracking\else \shortstack{\strut Object Tracking\\[-2pt]\rule{58pt}{0.4pt}}\fi} & \multicolumn{2}{>{\columncolor{tSeenHead}}c}{\iflatexml Hand Tracking\else \shortstack{\strut Hand Tracking\\[-2pt]\rule{58pt}{0.4pt}}\fi} & \cellcolor{tUnseenHead} & \multicolumn{2}{>{\columncolor{tUnseenHead}}c}{\iflatexml Object Tracking\else \shortstack{\strut Object Tracking\\[-2pt]\rule{58pt}{0.4pt}}\fi} & \multicolumn{2}{>{\columncolor{tUnseenHead}}c}{\iflatexml Hand Tracking\else \shortstack{\strut Hand Tracking\\[-2pt]\rule{58pt}{0.4pt}}\fi} \\
\iflatexml Method\else \multirow{-3}{*}{Method}\fi & \cellcolor{tSeenHead} \iflatexml \shortstack{SR\\(\%) $\uparrow$}\else \multirow{-2}{*}{\shortstack{SR\\(\%) $\uparrow$}}\fi & \cellcolor{tSeenHead} \shortstack{$E_\mathrm{pos}$\\(mm) $\downarrow$} & \cellcolor{tSeenHead} \shortstack{$E_\mathrm{rot}$\\($^\circ$) $\downarrow$} & \cellcolor{tSeenHead} \shortstack{$E_\mathrm{tip}$\\(mm) $\downarrow$} & \cellcolor{tSeenHead} \shortstack{$E_\mathrm{joint}$\\($^\circ$) $\downarrow$} & \cellcolor{tUnseenHead} \iflatexml \shortstack{SR\\(\%) $\uparrow$}\else \multirow{-2}{*}{\shortstack{SR\\(\%) $\uparrow$}}\fi & \cellcolor{tUnseenHead} \shortstack{$E_\mathrm{pos}$\\(mm) $\downarrow$} & \cellcolor{tUnseenHead} \shortstack{$E_\mathrm{rot}$\\($^\circ$) $\downarrow$} & \cellcolor{tUnseenHead} \shortstack{$E_\mathrm{tip}$\\(mm) $\downarrow$} & \cellcolor{tUnseenHead} \shortstack{$E_\mathrm{joint}$\\($^\circ$) $\downarrow$} \\
\midrule
Retarget replay & $0.0$ & $20.44$ & $14.81$ & $\mathbf{1.88}$ & $\mathbf{0.71}$ & $0.0$ & $17.23$ & $14.69$ & $\mathbf{1.98}$ & $\mathbf{0.68}$ \\
SPIDER$^{\dagger}$~\cite{spider} & $9.7{\scriptstyle \pm 0.5}$ & $21.97$ & $\mathbf{6.38}$ & $10.75$ & $5.53$ & $0.5{\scriptstyle \pm 0.5}$ & $18.31$ & $\mathbf{5.94}$ & $8.47$ & $4.27$ \\
\mbox{ConTrack (specialist)$^{\dagger}$\,\cite{contrack}} & $49.8{\scriptstyle \pm 1.0}$ & $17.56$ & $10.79$ & $14.76$ & $9.34$ & $32.7{\scriptstyle \pm 1.8}$ & $16.22$ & $10.19$ & $15.13$ & $9.07$ \\
SimToolReal (fine-tuned)$^\S$~\cite{simtoolreal} & $32.6{\scriptstyle \pm 1.3}$ & $20.70$ & $24.66$ & $\text{---}$ & $\text{---}$ & $9.5{\scriptstyle \pm 1.4}$ & $20.85$ & $23.34$ & $\text{---}$ & $\text{---}$ \\
DexTrack~\cite{dextrack} & $14.2{\scriptstyle \pm 0.8}$ & $13.45$ & $18.43$ & $12.44$ & $8.41$ & $0.0{\scriptstyle \pm 0.0}$ & $15.47$ & $17.04$ & $14.12$ & $8.91$ \\
\midrule
\textbf{\DMAColor{} (ours)} & \cellcolor{tSeenOurs} $\mathbf{80.4}{\scriptstyle \boldsymbol{\pm 0.9}}$ & \cellcolor{tSeenOurs} $\mathbf{9.14}$ & \cellcolor{tSeenOurs} $8.30$ & \cellcolor{tSeenOurs} $8.71$ & \cellcolor{tSeenOurs} $5.32$ & \cellcolor{tUnseenOurs} $\mathbf{69.7}{\scriptstyle \boldsymbol{\pm 2.1}}$ & \cellcolor{tUnseenOurs} $\mathbf{9.02}$ & \cellcolor{tUnseenOurs} $8.41$ & \cellcolor{tUnseenOurs} $6.38$ & \cellcolor{tUnseenOurs} $4.25$ \\
\bottomrule
\end{tabularx}
\par\smallskip
  {\raggedright\footnotesize Success: object pose error within 50\,mm and $45^\circ$ at every reference frame. SR: mean $\pm$ sample SD; errors: mean.{} Bold: best in each column. $^\dagger$Optimized or trained on each evaluation reference.
   $^\S$Uses reference waypoints sampled every 0.5\,s (same thresholds; 10\,s per waypoint); errors are averaged over reached waypoints; checkpoint selected by SR on the evaluation set; SD over five evaluation orders.\par}
\end{table*}

\subsection{Policy Learning through a Hand--Object World Model}
\label{sec:learning}
One policy for all DexEmpla references converges slowly with PPO~\cite{ppo} (Sec.~\ref{sec:exp_ablation}), and MBRL can speed it up.
SuperMimic~\cite{supermimic} explicitly learns one differentiable world model of the body dynamics shared by all references, which greatly shortens training.
We want the same efficiency for hand--object tracking.
\DMA{} fits a hand--object world model to $K_w$-step windows of logged simulator transitions.
The policy is then rolled out inside the model for $H$ steps.
The return of these imagined rollouts, bootstrapped by a critic, is backpropagated through the model to the policy ($K_w$ and $H$ in Appendix~\ref{app:train}).

\textbf{World model.}
The state $s_t$ contains the finger joint angles and velocities and the object pose and velocities, all in the current wrist frame.
The model takes the executed finger targets $u_t$, the reference wrist motion $m_t$, the force and torque $d_t$ applied to the object, and a context $\zeta$.
The context holds the randomized physical parameters and an object embedding.
The model predicts the next state and the contact probability:
\begin{equation}
\hat s_{t+1}=\hat s_t\oplus f_\omega(\hat s_t,u_t,m_t,d_t,\zeta),\quad
\hat c_{t+1}=\sigma\big(c_\omega(\hat s_{t+1},\zeta)\big),
\label{eq:wm}
\end{equation}
where $f_\omega$ outputs a state increment and $\sigma$ is the logistic function.
Compared with SuperMimic, our model adds an object embedding and a contact head.

\textit{Object embedding.}
The world model and the critic are used only in training, where every object is known, so they need only distinguish between the training objects.
The world model therefore takes a learned embedding of each training object in place of the actor's local surface points, which avoids selecting and encoding surface points at every step of the imagined rollouts (Sec.~\ref{sec:exp_ablation}); the critic receives the same embedding as a privileged input.

\textit{Contact head.}
The reward of \DMA{} depends on contact, so we use a head $c_\omega$ to predict the probability $\hat c$ that at least two fingers touch the object ($n^{c}\ge 2$).
In imagination the gate \eqref{eq:gate} becomes $\hat\chi_t=h_{k_t}\hat c_t$.
The object reward is then computed from predicted states, and its gradient reaches the actions that make or break the grasp.

\textbf{Domain randomization.}
During training, we vary hand friction, object mass and center of mass, PD gains, and joint offsets.
We add noise to object pose observations, simulate marker occlusion, and randomly mask actions~\cite{teledexter}.
To improve robustness to unexpected object motion, we apply random forces and torques to the object during hand--object interaction.
Appendix~\ref{app:dr} lists the ranges.

\subsection{Reset Sampling with Segment Exclusion}
\label{sec:exclusion}

Motion tracking methods start each episode from a reference state~\cite{deepmimic} and prioritize segments that the policy tracks poorly~\cite{supermimic}.
However, some segments are physically infeasible because hand--object references suffer from capture and retargeting errors.
The policy's return on such a segment stays low no matter how long training runs, so prioritized sampling keeps choosing it at the expense of the other segments.

Infeasible segments are hard to identify in advance.
\DMA{} therefore excludes segments where episodes keep failing.
We cut the training references of all objects into one pool of 1\,s segments, choose the start segment by return-based reset sampling~\cite{supermimic}, and start the episode from a frame drawn uniformly within it.
Each failure is attributed to the segment in which the episode terminates, and for each segment we track a visit count $n_j$ and a smoothed failure rate $F_j$.
A segment is permanently excluded once $n_j\ge n_{\min}$ and $F_j\ge F_{\max}$.
At most a fraction $\epsilon$ of all segments can be excluded (values in Appendix~\ref{app:train}), and a reference is dropped once half of its segments are.

\section{Experiments}
\suppressfloats[t]
Our experiments evaluate \DMA{} through three questions:
\begin{enumerate}
  \renewcommand{\labelenumi}{Q\arabic{enumi}.}
  \item Does \DMA{} generalize in-hand skills to unseen objects and references better than prior methods?
  \item Can \DMA{} perform real-world tool use with unseen tools, both in motion tracking and in teleoperation?
  \item How do \DMA{}'s design choices and training data size affect policy performance and training efficiency?
\end{enumerate}

\setcounter{dbltopnumber}{1}
\begin{figure*}[t]
  \centering
  \includegraphics[width=\textwidth]{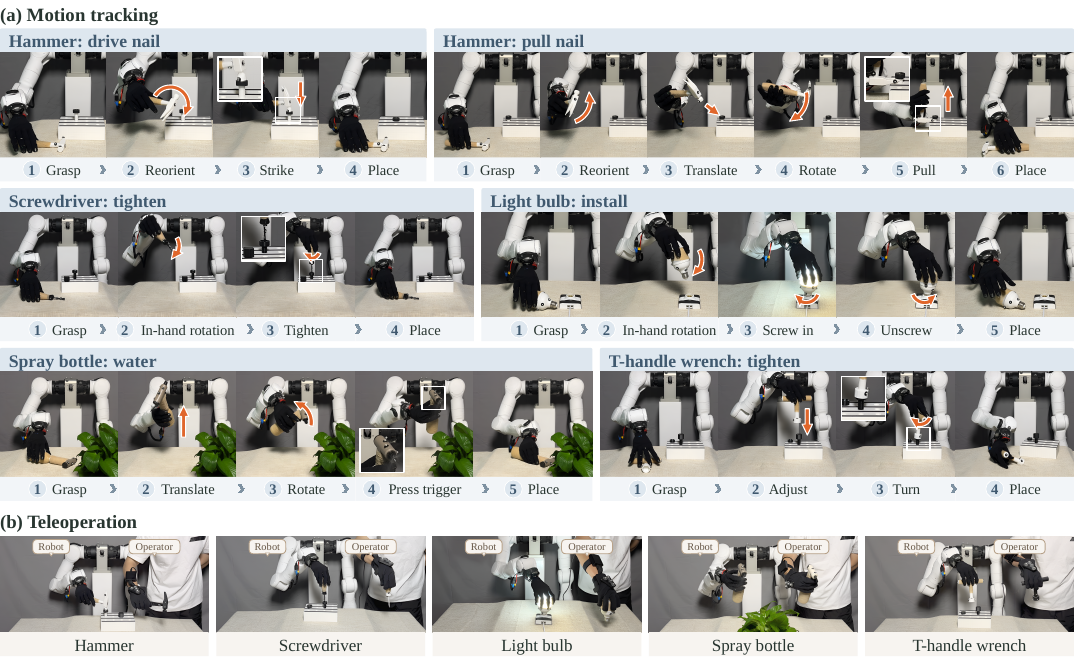}
  \caption{\textbf{Real-world tool use with unseen tools.} (a) Motion tracking, with the stages of each task. (b) Teleoperation.}
  \label{fig:realstage}
\end{figure*}

\subsection{Experimental Setup}
\label{sec:exp_setup}
\textbf{Simulation setup.}
We simulate in MuJoCo Warp at a 50\,Hz control rate.

\textbf{Real-world platform.}
A 20-DoF Wuji Hand~2 is mounted on a 7-DoF Tianji arm, and an optical motion-capture system provides the wrist and object poses.
The arm tracks the reference wrist pose through inverse kinematics and joint impedance control.

\subsection{Comparison with Prior Methods}
\label{sec:exp_cmp}
To answer Q1, we compare \DMA{} with five baselines on unseen references of \emph{seen objects} (315~references) and \emph{unseen objects} (120~references across 9~tools); both sets cover rotation, translation, and combined motions (Table~\ref{tab:data}).

\textbf{Baselines.}
We compare with \emph{Retarget replay}, which executes the retargeted joint trajectories open loop, and four prior methods with public code: SPIDER~\cite{spider}, a ConTrack-style specialist~\cite{contrack}, DexTrack~\cite{dextrack}, and SimToolReal~\cite{simtoolreal}.
All four build on their official implementations.
SPIDER runs with its official settings without contact guidance, as in its large-scale data generation.
The specialist trains one policy per reference.
DexTrack is retrained on our training set.
These three are adapted to the Wuji Hand~2 model used by \DMA{}.
We fine-tune SimToolReal from its publicly released checkpoint on our 34 training object geometries, so it keeps its original training objective and native KUKA--SHARPA embodiment.

\textbf{Metrics.}
We evaluate without dynamics randomization.
We report the success rate and the mean errors of object pose ($E_\text{pos}$, $E_\text{rot}$), fingertip position in the palm frame ($E_\text{tip}$), and finger joint angle ($E_\text{joint}$).
Errors are averaged over the interaction frames before each rollout fails or ends.

\textbf{Overall results.}
\DMA{} achieves the highest SR on both splits (Table~\ref{tab:sim}) and exceeds the ConTrack specialists trained on the evaluation references by 30.6 and 37.0 percentage points, which demonstrates its generalization.
DexTrack, retrained on the same data, completes 14.2\% and 0.0\%, so in-hand demonstrations alone do not give a generalist tracker these skills.
The object-centric SimToolReal is the stronger generalist baseline on seen objects (32.6\%) but retains only 29\% of this SR on unseen objects, whereas \DMA{} retains 87\%.
Without human hand references, it adopts unnatural finger postures~\cite{adadex} and cannot follow the reference object motion.
The two retargeting baselines, replay and SPIDER, output an open-loop control sequence per reference, which cannot adjust the fingers reactively, and complete 0.0\% and at most 9.7\%.
In-hand manipulation thus needs a reactive controller.

For tracking errors, replay follows the hand reference but rarely establishes the intended interaction, so it has the lowest hand errors and large object errors.
\DMA{} achieves low tracking errors, although the references contain capture and retargeting errors that it must correct to interact with the object.

\DMA{} fails mostly on two kinds of references.
The first involves objects with unusual geometry in the long tail of the data, such as the 12-mm-thick plates (30.0\% and 49.5\%).
The second relies on wrist coordination and object inertia, such as pivoting a bar; the policy controls only the fingers and cannot adjust the wrist.

\subsection{Real-World Evaluation}
\label{sec:exp_real}
\begin{table}[t]
  \centering
  \caption{Real-world tool use with unseen tools.}
  \label{tab:real}
  \setlength{\tabcolsep}{1.5pt}
  \renewcommand{\arraystretch}{1.05}
  \footnotesize
  \renewcommand{\tabularxcolumn}[1]{m{#1}}
  \begin{tabularx}{\columnwidth}{
    >{\hsize=1.96\hsize\linewidth=\hsize\centering\arraybackslash}X
    *{2}{>{\columncolor{realTrack}\hsize=0.76\hsize\linewidth=\hsize\centering\arraybackslash}X}*{2}{>{\columncolor{realTele}\hsize=0.76\hsize\linewidth=\hsize\centering\arraybackslash}X}}
    \toprule
    & \multicolumn{2}{>{\columncolor{realTrackHead}}c}{Motion tracking}
    & \multicolumn{2}{>{\columncolor{realTeleHead}}c}{Teleoperation} \\
    \cmidrule(lr){2-3}\cmidrule(lr){4-5}
    Task & \cellcolor{realTrackHead}SR (\%) $\uparrow$ & \cellcolor{realTrackHead}TP (\%) $\uparrow$ & \cellcolor{realTeleHead}SR (\%) $\uparrow$ & \cellcolor{realTeleHead}TP (\%) $\uparrow$ \\
    \midrule
    Hammer: drive nail   & 86.7 & 93.3 & 93.3 & 98.3 \\
    Hammer: pull nail    & 66.7 & 82.2 & 73.3 & 85.6 \\
    Screwdriver: tighten & 86.7 & 90.0 & 93.3 & 96.7 \\
    Light bulb: install  & 73.3 & 84.0 & 73.3 & 86.7 \\
    Spray bottle: water  & 60.0 & 80.0 & 80.0 & 90.7 \\
    T-handle wrench: tighten    & 53.3 & 71.7 & 66.7 & 83.3 \\
    \midrule
    Mean                 & 71.1 & 83.5 & 80.0 & 90.2 \\
    \bottomrule
  \end{tabularx}
  \par\vspace{2pt}
  {\footnotesize\raggedright Success: all stages (Fig.~\ref{fig:realstage}(a)) completed without dropping the tool. TP: task progress, the fraction of completed stages. 15 trials per task and mode, without retries; in motion tracking, five references, each executed three times.\par}
\end{table}

\begin{table}[t]
  \centering
  \caption{Ablations of \DMA{} components.}
  \label{tab:ablation_summary}
  \setlength{\tabcolsep}{3pt}
  \renewcommand{\arraystretch}{1.3}
  \footnotesize
  \definecolor{seenBg}{HTML}{F1F7F3}
  \definecolor{unseenBg}{HTML}{F7F3FB}
  \definecolor{seenHead}{HTML}{DCECE2}
  \definecolor{unseenHead}{HTML}{E8DEF2}
  \definecolor{seenOurs}{HTML}{DCEDE2}
  \definecolor{unseenOurs}{HTML}{EADDF5}
  \setlength{\aboverulesep}{0pt}
  \setlength{\belowrulesep}{0pt}
  \setlength{\cmidrulesep}{0pt}
  \begin{tabularx}{\linewidth}{@{}l>{\raggedright\arraybackslash}X>{\columncolor{seenBg}}c>{\columncolor{unseenBg}[\tabcolsep][0pt]}c@{}}
    \toprule
    Component & Variant & \cellcolor{seenHead}Seen SR & \cellcolor{unseenHead}Unseen SR \\
    \midrule
    \multirow{4}{*}{Representation}
      & TopoRetarget~\cite{toporetarget} (Axis) & $72.7\pm1.3$ & $52.6\pm3.9$ \\
      & Dex4D~\cite{dex4d} (Global64) & $77.7\pm1.0$ & $41.4\pm0.8$ \\
      & AdaDex~\cite{adadex} (Pose+BPS) & $74.2\pm1.8$ & $26.9\pm0.8$ \\
      & Dexplore~\cite{dexplore} (Pose+vectors) & $78.6\pm0.9$ & $59.2\pm6.4$ \\
    \hline
    \multirow{2}{*}{Sampling}
      & Uniform & $76.4\pm1.0$ & $62.7\pm5.7$ \\
      & Return-based & $78.3\pm0.9$ & $65.1\pm2.5$ \\
    \hline
    \multirow{2}{*}{Reward gate}
      & DexPoint~\cite{dexpoint} (Contact-aware) & $74.9\pm1.4$ & $60.7\pm6.6$ \\
      & DexTrack~\cite{dextrack} (Always-on) & $78.3\pm1.1$ & $64.9\pm5.4$ \\
    \midrule
    Full model & \textbf{Ours} & \cellcolor{seenOurs}$\mathbf{80.4\pm0.9}$ & \cellcolor{unseenOurs}$\mathbf{69.7\pm2.1}$ \\
    \bottomrule
  \end{tabularx}
  \par\vspace{2pt}
  {\footnotesize\raggedright Each variant replaces one DMA component. Work names identify the related prior designs. SR (\%): mean $\pm$ sample SD over three training seeds.\par}
\end{table}
\begin{figure*}[t]
  \centering
  \begin{minipage}[t]{0.364\textwidth}
    \centering
    \includegraphics[width=\linewidth]{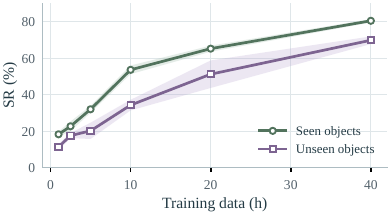}\\[-1pt]
    {\footnotesize (a)}
  \end{minipage}\hfill
  \begin{minipage}[t]{0.238\textwidth}
    \centering
    \includegraphics[width=\linewidth]{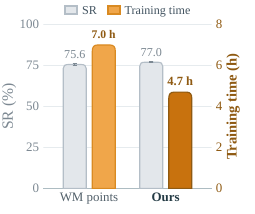}\\[-1pt]
    {\footnotesize (b)}
  \end{minipage}\hfill
  \begin{minipage}[t]{0.364\textwidth}
    \centering
    \includegraphics[width=\linewidth]{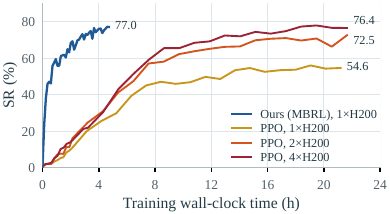}\\[-1pt]
    {\footnotesize (c)}
  \end{minipage}
  \caption{\textbf{Data size, world-model input, and training efficiency.} (a) SR versus training data; lines: mean, bands: $\pm1$ sample SD over three training seeds. (b) World-model (WM) input, with local surface points (WM points) in place of the object embedding (Ours): SR on all evaluation references and training time; error bars: SD over five evaluation seeds; training time excludes evaluation. (c) SR on all evaluation references versus training wall-clock time.}
  \label{fig:ablation_row}
\end{figure*}

To answer Q2, we deploy the \DMA{} policy on six tasks with unseen tools in the real-world setup (Table~\ref{tab:real}).
In motion tracking (Fig.~\ref{fig:realstage}(a)), the policy tracks a recorded human demonstration.
In teleoperation (Fig.~\ref{fig:realstage}(b)), the operator wears the Wuji Glove and manipulates a tracked tool while the robot operates an identical copy.

\textbf{Results.}
\DMA{} is trained in simulation on geometric primitives and deployed without fine-tuning on unseen everyday tools.
It completes 71.1\% of the trials in motion tracking and 80.0\% in teleoperation (Table~\ref{tab:real}).
Each task chains several skills; nail pulling, for example, involves pick-and-place, in-hand rotation and translation, and regrasping.
One policy thus executes in-hand skills and combines them within a task.
The T-handle wrench has the lowest SR, as it must engage a screw that the policy does not observe.
Teleoperation has a higher SR because the operator can adjust the reference online, for example to align the tool.

\subsection{Ablation Studies}
\label{sec:exp_ablation}

To answer Q3, we ablate \DMA{}'s components (Table~\ref{tab:ablation_summary}), training data size, and training algorithm.

\textbf{Object representation.}
We compare with four alternatives.
\emph{Axis}~\cite{toporetarget} gives only the object pose.
\emph{Global64}~\cite{dex4d} and \emph{Pose+BPS}~\cite{adadex} describe the whole object, as point pairs over the mesh and as a basis point set.
\emph{Pose+vectors}~\cite{dexplore} describes the surface near the hand, as link-to-surface vectors.
Seen-object SR varies little (72.7--80.4\%) and is lowest for \emph{Axis}, which sees no geometry.
On unseen objects, whole-object geometry falls below the pose-only \emph{Axis} (41.4\% and 26.9\% vs.\ 52.6\%), whereas geometry near the hand rises above it (59.2\%, and 69.7\% for our representation).
The unseen objects differ from the training primitives in shape, but local geometry near the contact area differs less~\cite{dexrepnet}.
Our representation shares its point pairs and encoder with \emph{Global64} and differs only in which 64 points are selected, which alone raises unseen-object SR from 41.4\% to 69.7\%.
Representing local contact geometry is thus crucial for generalization.

\textbf{Reset sampling.}
Prioritizing low-return segments (\emph{Return-based}) improves on \emph{Uniform}~\cite{deepmimic} by 1.9 and 2.4 points on seen and unseen objects.
The full model additionally excludes persistently failing segments, which adds a further 2.1 and 4.6 points.

Manual inspection shows that the excluded segments contain physically infeasible references, such as severe hand--object penetration.
These segments have a low return but are poor initial states.
Excluding them means that the policy reaches them only from nearby feasible initial states, which improves training.

\textbf{Reward gate.}
\emph{Contact-aware} ($\chi_t=\mathbb{I}[n^c_t\ge 2]$, omitting $h_{k_t}$) performs worst, 5.5 and 9.0 points below the full model.
It activates the object reward once two fingers touch the object, even while the object rests on the table.
The policy can hack this by touching the object early during approach and lingering during release.
\emph{Always-on} ($\chi_t=1$) pays the object reward while the object rests untouched, so it does not encourage interaction, and is 2.1 and 4.8 points below the full model.

Our gate is the most effective: with the interaction label, it pays the object reward only when interaction should occur, and it still encourages contact.

\textbf{Dataset size.}
We train on nested 1, 2.5, 5, 10, and 20\,h subsets of whole capture sessions; each contains all 34 training objects, so gains come from more interaction data, not more objects (Fig.~\ref{fig:ablation_row}(a)).
SR increases monotonically with data, from 18.3\% to 80.4\% on seen objects and from 11.4\% to 69.7\% on unseen objects.
More data thus improves generalization, without saturation at 40.1\,h.
Scaling up in-hand demonstrations is a promising way to improve it further.

\textbf{World-model input.}
We compare the embedding with local surface points computed from the predicted object pose (WM points, Fig.~\ref{fig:ablation_row}(b)).
The embedding reaches a similar SR (77.0\% vs.\ 75.6\%) and reduces training time from 7.0 to 4.7\,h.
It thus speeds up training without reducing SR, indicating that the embedding gives the world model enough object information for policy learning.

\textbf{Training efficiency.}
We replace MBRL with PPO~\cite{ppo} on the same references, inputs, and seed (Fig.~\ref{fig:ablation_row}(c)).
PPO matches MBRL's SR only on 4$\times$H200 after about 18\,h, using about 15 times the GPU-hours; MBRL thus substantially accelerates training.

\section{Conclusion}
We presented \textbf{DexMimicAnything} (\DMA{}), a hand--object co-tracking controller, and \textbf{DexEmpla}, a human demonstration dataset focused on in-hand manipulation.
\DMA{} combines a representation of local contact geometry, interaction-aware object tracking, and segment exclusion in a model-based RL framework with a differentiable world model.
It outperforms the evaluated baselines and transfers to real-world tool use without fine-tuning.

The current system relies on optical motion capture, which constrains both data collection and deployment.
Future work will explore vision-based hand--object state estimation to remove this dependence.
Thin plates and motions that require wrist coordination remain the hardest cases.
As generalization has not saturated with data, we will also scale up the demonstrations.\looseness=-1

\section*{Appendix}
\setcounter{subsection}{0}
\renewcommand{\thesubsection}{\Alph{subsection}}
\subsection{Interaction Labels, Surface Points, and Reward}
\label{app:method}
Table~\ref{tab:hyper} lists the parameters of the interaction labels (Sec.~\ref{sec:data}), the local surface point pairs (Sec.~\ref{sec:repr}), and the reward \eqref{eq:reward}.
The reward terms are
\begin{equation}
\begin{aligned}
r^{\mathrm{obj}}_t&=w_p\phi^{p}_t+w_a\phi^{a}_t,\\
r^{\mathrm{hand}}_t&=w_l\phi^{l}_t+[w_q+w'_q(1-h_{k_t})]\phi^{q}_t,\\
r^{\mathrm{reg}}_t&=w_c c^{a}_t+w_s n^{sc}_t.
\end{aligned}
\label{eq:reward_terms}
\end{equation}
Each tracking term is a kernel $\phi^{x}_t=\exp(-(e^{x}_t/\sigma_x)^2)$: $e^{p}_t$ is the object position error in the wrist frame and $e^{a}_t$ the mean error of the six object keypoints, while $\phi^{l}_t$ and $\phi^{q}_t$ average the kernel over the link position errors and the range-normalized joint errors.
The penalty $c^{a}_t$ sums the squared first- and second-order action differences, and $n^{sc}_t$ counts finger self-collisions.

\iflatexml%
\begin{table}%
\else%
\par\medskip\noindent\begin{minipage}{\columnwidth}
\setcaptype{table}%
\fi
\centering
\footnotesize
\renewcommand{\arraystretch}{0.9}
\caption{Method parameters.}
\label{tab:hyper}
\setlength{\tabcolsep}{4pt}
\begin{tabular}{l l l}
  \toprule
  Group & Parameter & Value \\
  \midrule
  Interaction & gravity-test duration & 0.5\,s \\
  labels & displacement threshold $\delta_p$ & 5--10\,mm (per reference) \\
  & rotation threshold $\delta_R$ & $5^\circ$ \\
  \midrule
  Surface & surface points per object $N_b$ & $2{,}048$ \\
  points & selected points per hand $K$ & $32$ \\
  \midrule
  Reward & kernel widths $\sigma_p/\sigma_a/\sigma_l/\sigma_q$ & 0.05\,/\,0.04\,/\,0.025\,m, 0.1 \\
  & object weights $w_p/w_a$ & $2/6$ \\
  & hand weights $w_l/w_q/w'_q$ & $3/3/3$ \\
  & regularization weights $w_c/w_s$ & $0.1/1$ \\
  \bottomrule
\end{tabular}
\iflatexml%
\end{table}%
\else%
\end{minipage}\par\medskip%
\fi

\subsection{Policy Training}
\label{app:train}
Table~\ref{tab:train} lists the network widths and training hyperparameters.
All runs use 16{,}384 environments on one H200 GPU.

\iflatexml%
\begin{table}%
\else%
\par\medskip\noindent\begin{minipage}{\columnwidth}
\setcaptype{table}%
\fi
\centering
\footnotesize
\renewcommand{\arraystretch}{0.9}
\caption{Network architectures and training hyperparameters.}
\label{tab:train}
\setlength{\tabcolsep}{4pt}
\begin{tabular}{l l l}
  \toprule
  Group & Hyperparameter & Value \\
  \midrule
  Networks & actor MLP & [1024, 512, 256] \\
  & critic MLP & [1024, 1024, 512, 256] \\
  & PointNet & [64, 64, 128] \\
  & world-model MLP & [2048, 2048, 2048] \\
  \midrule
  World & window length $K_w$ & $8$ \\
  model & imagination horizon $H$ & $32$ \\
  & discount $\gamma$ & $0.99$ \\
  & TD parameter $\lambda$ & $0.95$ \\
  \midrule
  Optimization & parallel environments & $16{,}384$ \\
  & training iterations & 40k \\
  & optimizer & Adam \\
  & learning rate (cosine) & $10^{-3}\rightarrow3\times10^{-5}$ \\
  & gradient-norm clip & $1.0$ \\
  \midrule
  Reset & segment length & 1\,s \\
  sampling & reset sampling & as in~\cite{supermimic} \\
  & exclusion: min.\ visits $n_{\min}$ & $20$ \\
  & exclusion: failure rate $F_{\max}$ & $0.9$ \\
  & exclusion: max.\ fraction $\epsilon$ & 5\% \\
  \bottomrule
\end{tabular}
\iflatexml%
\end{table}%
\else%
\end{minipage}\par\medskip%
\fi

\subsection{Domain Randomization}
\label{app:dr}
Table~\ref{tab:dr} lists the randomization ranges ($U$: uniform; $\mathrm{LogU}$: log-uniform; $\pm x$: $U(-x,x)$).
Dynamics parameters are sampled once per environment, while observation noise, marker occlusion, and action masking are resampled at each control step.
Random forces, action masking, and marker occlusion ramp up linearly from iteration 2{,}000 to 12{,}000.

\iflatexml%
\begin{table}%
\else%
\par\medskip\noindent\begin{minipage}{\columnwidth}
\setcaptype{table}%
\fi
\centering
\footnotesize
\renewcommand{\arraystretch}{0.9}
\caption{Domain-randomization ranges.}
\label{tab:dr}
\setlength{\tabcolsep}{4pt}
\begin{tabular}{l l l}
  \toprule
  Group & Parameter & Range \\
  \midrule
  Hand & friction & $U(0.2, 1.0)$ \\
  & $k_p$ (scale) & $\mathrm{LogU}(0.75, 1.5)$ \\
  & $k_d$ (scale) & $\mathrm{LogU}(0.5, 2.0)$ \\
  & joint offset & $\pm0.01$\,rad \\
  Object & mass, inertia (scale) & $U(0.5, 2.0)$ \\
  & center of mass ($z$) & $\pm20\%$ \\
  & initial pose (per axis) & $\pm5$\,mm, $\pm15^\circ$ \\
  \midrule
  Object wrench & force, torque (per axis) & $\pm4$\,N, $\pm0.04$\,N$\cdot$m \\
  & duration, interval & 0.02\,s, $U(0.5, 1.2)$\,s \\
  Action masking & probability & $\le0.15$ \\
  & joints, duration & 3, $\le10$ steps \\
  \midrule
  Observation noise & joint angle & $\pm0.015$\,rad \\
  & joint velocity & $\pm0.05$\,rad/s \\
  & object position (per axis) & $\pm8$\,mm \\
  & object orientation & $\pm0.05$\,rad \\
  & marker occlusion probability & $\le0.15$ \\
  \bottomrule
\end{tabular}
\iflatexml%
\end{table}%
\else%
\end{minipage}\par\medskip%
\fi

\bibliographystyle{IEEEtran}
\bibliography{references}

\end{document}